\documentclass[letterpaper]{article} 
\usepackage{aaai2026}  
\usepackage{times}  
\usepackage{helvet}  
\usepackage{courier}  
\usepackage[hyphens]{url}  
\usepackage{graphicx} 
\usepackage{natbib}  
\usepackage{caption} 
\usepackage{algorithm}

\usepackage{subcaption}
\usepackage{algpseudocode} 
\usepackage{amsmath}
\usepackage{multirow}
\usepackage{amsfonts} 
\usepackage{newfloat}
\usepackage{listings}
\DeclareCaptionStyle{ruled}{labelfont=normalfont,labelsep=colon,strut=off} 
\floatstyle{ruled}
\newfloat{listing}{tb}{lst}{}
\floatname{listing}{Listing}
\nocopyright 

\title{SHeRL-FL: When Representation Learning Meets Split Learning in Hierarchical Federated Learning}

\author{\normalfont
    Dung T. Tran\textsuperscript{\rm 1},
    Nguyen B. Ha\textsuperscript{\rm 1},
    Van-Dinh Nguyen\textsuperscript{\rm 1,2},
    Kok-Seng Wong\textsuperscript{\rm 1,2}\thanks{Corresponding author.} \\
    \textsuperscript{\rm 1}Center for Environmental Intelligence, VinUniversity, Hanoi, Vietnam \\
    \textsuperscript{\rm 2}College of Engineering and Computer Science, VinUniversity, Hanoi, Vietnam \\
    \{dung.tt2, nguyen.hb, dinh.nv2, wong.ks\}@vinuni.edu.vn
}

\begin{document}

\maketitle

\begin{abstract}
    Federated learning (FL) is a promising approach for addressing scalability and latency issues in large-scale networks by enabling collaborative model training without requiring the sharing of raw data. However, existing FL frameworks often overlook the computational heterogeneity of edge clients and the growing training burden on resource-limited devices. However, FL suffers from high communication costs and complex model aggregation, especially with large models. Previous works combine split learning (SL) and hierarchical FL (HierFL) to reduce device-side computation and improve scalability, but this introduces training complexity due to coordination across tiers. To address these issues, we propose SHeRL-FL, which integrates SL and hierarchical model aggregation and incorporates representation learning at intermediate layers. By allowing clients and edge servers to compute training objectives independently of the cloud, SHeRL-FL significantly reduces both coordination complexity and communication overhead. To evaluate the effectiveness and efficiency of SHeRL-FL, we performed experiments on image classification tasks using CIFAR-10, CIFAR-100, and HAM10000 with AlexNet, ResNet-18, and ResNet-50 in both IID and non-IID settings. In addition, we evaluate performance on image segmentation tasks using the ISIC-2018 dataset with a ResNet-50-based U-Net. Experimental results demonstrate that SHeRL-FL reduces data transmission by over 90\% compared to centralized FL and HierFL, and by 50\% compared to SplitFed, which is a hybrid of FL and SL, and further improves hierarchical split learning methods. 
\end{abstract}


\section{Introduction}

    Federated Learning (FL) \cite{konečný2015federatedoptimizationdistributedoptimizationdatacenter, mcmahan2017communication} enables multiple clients to collaboratively train a shared model without exchanging raw data, preserving privacy and data sovereignty in domains such as healthcare, finance, and mobile computing. Despite its advantages, large-scale FL suffers from high communication costs and heavy computation on resource-constrained clients. To alleviate these limitations, hierarchical architectures have been proposed to improve communication efficiency by introducing intermediate edge servers. However, even with hierarchical FL, the computational burden on clients remains a major challenge. To address this, Split Learning (SL) \cite{GUPTA20181} reduces client computation by partitioning a deep neural network between clients and a server. In SL, the client-side model processes local data and sends intermediate activations (smashed data) to the server, which completes the forward pass, computes the loss, and returns gradients for backpropagation. This approach leverages server computation to reduce client workload and has shown promise in various applications \cite{vepakomma2019split, poirot2019split, li2023convergence}. However, SL suffers from sequential bottlenecks, catastrophic forgetting \cite{doi:10.1073/pnas.1611835114} and prolonged client idle time, both of which degrade overall training efficiency.

     To combine the efficiency of SL and the scalability of FL, \cite{thapa2022splitfed} proposed SplitFed, which integrates model partitioning through parallel training. Each client maintains a local copy of the client-side partition, processes private data, and sends smashed data to a central server that hosts the server-side partition. The server completes the forward and backward passes, returning gradients to clients for local updates. SplitFed allows parallel client training, improving scalability while retaining SL’s reduced client computation. However, SplitFed suffers from high client–server communication and synchronization delays \cite{thapa2022splitfed}. Hierarchical Split Federated Learning (HSFL) \cite{10980018} extends this design by adding the edge tier to lower communication costs, relieve device-level constraints, and accelerate convergence. Yet, HSFL often splits models by device capability rather than layer semantics, potentially hindering representation consistency. Independent tier-wise aggregation can destabilize training, and frequent gradient exchanges raise privacy concerns, such as gradient inversion attacks \cite{10025466, guo2025exploringvulnerabilitiesfederatedlearning}.

     \textbf{Motivation.} While recent FL frameworks improve scalability and communication, the quality of intermediate representations exchanged between tiers remains overlooked. Poorly structured features disrupt consistency, slow convergence, and degrade generalization. Representation learning can improve semantic alignment across segmented models, improve information flow, and stabilize aggregation. However, current split-based methods pay limited attention to updating intermediate features, as they are not treated as explicit learning objectives. In this paper, we propose \textbf{SHeRL-FL} (Split and Hierarchical Representation Learning), a role-aware FL architecture that integrates explicit representation learning at edge servers. By aligning model segments with their functional roles, SHeRL-FL enables tier-wise optimization without backward gradients from the cloud, reducing communication and privacy risk.

    \textbf{Contribution.} In summary, the main contributions are as follows:
    \begin{itemize}
        \item We propose SHeRL-FL, a split hierarchical federated learning framework that integrates functional alignment into each tier. Unlike prior approaches that treat tiers uniformly or require end-to-end gradients, SHeRL-FL enables each tier to learn task-relevant representations independently while preserving inter-tier coherence.
        \item By assigning distinct objectives to client, edge, and cloud models, SHeRL-FL improves stability, interpretability, and modularity, making it well-suited for resource-constrained and privacy-sensitive settings.
        \item SHeRL-FL eliminates cloud-to-client gradient transmission, reducing communication costs and mitigating gradient-based privacy risks, which is critical for domains like healthcare and personal analytics.
        \item Experiments on CIFAR-10 ~\cite{krizhevsky2009learning}, CIFAR-100 ~\cite{krizhevsky2009learning}, HAM10000 ~\cite{tschandl2018ham10000}, and ISIC-2018 ~\cite{8363547} across multiple hierarchical settings show consistent gains in accuracy, efficiency, and representation quality over baselines such as FedAvg, FedSGD, FedNova, FedProx, SplitFed, HierFL, and HSFL.
    \end{itemize}

\section{Background and Related Works}

    Conventional FL algorithms include Federated Stochastic Gradient Descent (FedSGD) and Federated Averaging (FedAvg) \cite{mcmahan2017communication}. FedSGD aggregates after each mini-batch, causing frequent communication, while FedAvg improves efficiency by allowing multiple local updates before aggregation, better handling non-IID data. To address heterogeneity, FedProx \cite{MLSYS2020_1f5fe839} adds a proximal term for stable convergence. In addition, FedNova \cite{10.5555/3495724.3496362} tackles client drift in heterogeneous systems by normalizing local updates. 
 
     SL introduced by \cite{GUPTA20181, vepakomma2019split} partitions deep neural networks $\mathbf{W}$ between clients and a server. Clients send smashed data, rather than raw data, to the server, reducing model leakage \cite{9756883}. \cite{10.1145/3320269.3384740} proposed Differential Privacy (DP) incorporated into SL, integrating privacy guarantees while maintaining the utility of the model. However, traditional SL clients follow sequential training processes, resulting in a bottleneck on the server side, particularly on a large scale. SplitFed \cite{thapa2022splitfed} addresses this by allowing clients to train in parallel and send activations to a server, which trains the remaining model layers. Client models are periodically aggregated to maintain synchronized learning. Other variants further improve privacy, scalability, and latency, including SplitNN with partial sharing \cite{vepakomma2019split} and Asynchronous Federated SL \cite{10707222}, and domain-specific applications in healthcare \cite{10746501}, IoT anomaly detection \cite{bdcc8030021}, and cloud-edge recommendation systems \cite{10.1186/s13677-023-00435-5}, demonstrating its privacy-preserving benefits and adaptability across domains.

     To reduce cloud communication at scale, Hierarchical Federated Learning (HierFL) introduces intermediate aggregators (e.g., edge servers)\cite{9148862}. Clients send updates to edge nodes that perform partial aggregation before being sent to the cloud for final aggregation. This hierarchy reduces upstream communication and distributes the server-side computation. Extensions of this work include Hierarchical Global Asynchronous Federated Learning Across Multi-Center \cite{pmlr-v260-xie25c} with straggler-minimizing asynchronous aggregation and Hierarchical Clustering-based Personalized Federated Learning \cite{10.1145/3580795} for hierarchical clustering with non-IID personalization. However, existing HierFL frameworks overlook edge device heterogeneity and rely on client-side full-model training, which burdens constrained devices. This motivates the integration of computation-aware model splitting in a hierarchical structure. Hence, HSFL \cite{10980018} splits the model between clients, edge servers, and the cloud. \cite{10251444} further improves efficiency with distributed training and aggregation. However, most HSFL methods neglect functional tier-specific roles, leading to misaligned representations and suboptimal learning.

     Representation learning offers a potential solution to misalignment by guiding model segments to learn semantically meaningful features. Contrastive learning methods, including SimCLR \cite{10.5555/3524938.3525087} and SupCon \cite{10.5555/3495724.3497291}, have shown success in centralized settings, while ConPro \cite{Nguyen_2024_CVPR} and SemiSE \cite{10980996} propose role-conditioned and semi-supervised strategies for the representation of medical images. Although federated representation learning enhances generalization and personalization under non-IID data, its role in HSFL is still underexplored. Our work bridges this gap by integrating role-aware representation learning into HSFL to support more efficient, modular, and privacy-preserving training.

\section{The Proposed Framework}
    \begin{figure*}[!ht]
        \centering
        \includegraphics[width=\linewidth]{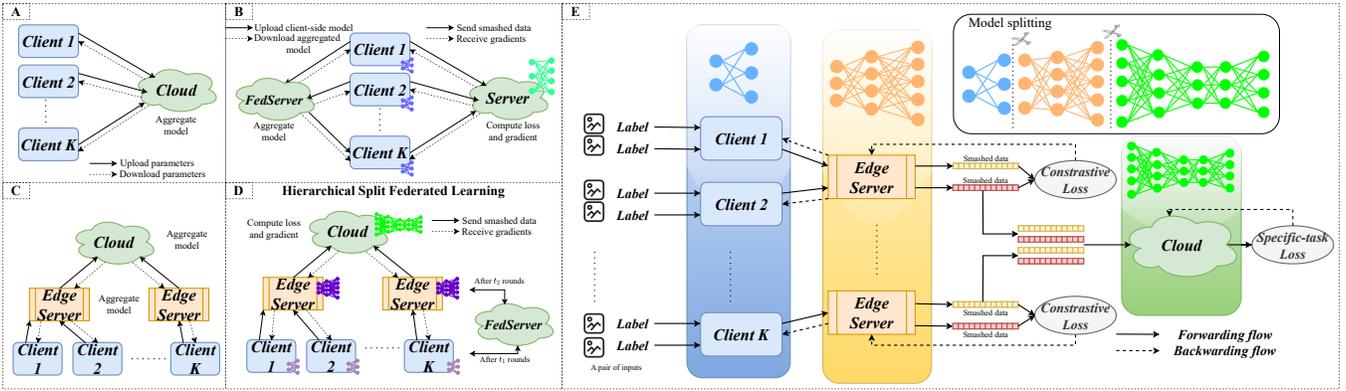}
        \caption{Overview of FL paradigms. (A) Conventional FL exchanges full models with the cloud. (B) SplitFL transmits smashed data but depends on cloud gradients. (C) Hierarchical FL uses edge servers for intermediate aggregation. (D) Hierarchical SplitFL combines model splitting with hierarchical aggregation. (E) SHeRL-FL (ours) enhances (D) with semantic-aware splitting, edge-level contrastive learning, and gradient-free communication to improve accuracy, efficiency, and scalability.}
        \label{fig:framework}
    \end{figure*}
    We propose \textit{Split and Hierarchical Representation Learning} (SHeRL-FL) to reduce communication and enhance representation alignment in federated systems. As shown in Figure~\ref{fig:framework} and Algorithm~\ref{alg:sherlfl}, SHeRL-FL operates in three tiers: clients, edge servers, and cloud. Training has three stages: (1) clients extract features and send smashed data to edges; (2) edges refine representations via contrastive learning and aggregate locally; (3) the cloud performs final inference and global aggregation.
    
     \textbf{Model Partitioning.} Unlike uniform partitioning in SL, our method aligns the splits with functional roles and system constraints. The client hosts shallow layers for low-level features (edges, textures, shapes), enabling lightweight, privacy-preserving computation. The edge server contains deeper layers for high-level representations and handles representation learning and local aggregation. The cloud holds the task-specific module (classifier/decoder) for final inference and global aggregation, ensuring cross-silo consistency. This function-aware design matches resources to roles, reducing redundancy and improving efficiency.
    
     \textbf{Training Workflow.} Each client processes raw input data using its local model segment and sends the smashed data output to its assigned edge server. Instead of processing these features independently, the edge server forms input pairs and applies contrastive learning to organize the feature space better. We adopt a margin-based contrastive loss that encourages similar samples to have similar representations and dissimilar samples to be far apart in the embedding space. Specifically, the loss is computed as follows:
    \begin{align}
        \mathcal{L}_{\text{contrastive}} &= \mathbb{E}\Big[ y_{ij} \, d_{\mathrm{cos}}(c_i, c_j) \\
        &\quad + (1 - y_{ij}) \, \max\big(0, m - d_{\mathrm{cos}}(c_i, c_j)\big) \Big] \nonumber
    \end{align}
    
    \begin{equation}
        d_{\text{cos}}(c_i, c_j) = \frac{c_i^T c_j}{\|c_i\| \|c_j\|},
    \end{equation}
     \noindent where $y_{ij} = 0$ if $c_i$ and $c_j$ belong to the same class, $y_{ij} = 1$ otherwise, and $m \in [0, 2]$ is a margin hyperparameter. The cosine distance encourages direction-based similarity, which is more suitable for representation learning compared to the Euclidean distance. The contrastive loss is backpropagated by the edges and partially into the clients, enabling immediate local updates without cloud interaction. This decoupling significantly reduces communication overhead and round-trip delays.
    
    \textbf{Hierarchical Model Aggregation.} At intervals $t_1$, each edge aggregates client updates via averaging or similar schemes and then forwards results to the cloud. The cloud computes task loss (e.g., cross-entropy $\mathcal{L}_{\text{task}}$) and performs global aggregation at $t_2$ across edges. This two-level process reduces client–cloud traffic, alleviates communication bottlenecks, enables faster edge updates, and optimizes representations locally before global alignment.

    \textbf{Advantages.} The framework offers key benefits for large-scale FL. Localized training and deferred cloud communication reduce bandwidth and latency. Clients send only intermediate features, improving privacy and reducing leakage risk. Edge-level contrastive learning improves feature space structure and aligns objectives within silos. Hierarchical aggregation balances workload across tiers, scaling to hundreds of clients. Local contrastive learning also boosts robustness under non-IID data by promoting intra-silo alignment.
    \begin{algorithm}[!ht]
        \caption{SHeRL-FL Training}
        \label{alg:sherlfl}
        \begin{algorithmic}[1]
            \State \textbf{Input:} Number of clients $N$, edge servers $E$, cloud server $C$, Number of epochs $E_p$, aggregation intervals $t_1, t_2$, client-side model $M_c$, edge-side model $M_e$, cloud-side model $M_{\mathrm{cloud}}$.
            \State \textbf{Output:} Trained $M_c, M_e, M_{\mathrm{cloud}}$
            \State Initialize $M_c^i, M_e^j, M_{\mathrm{cloud}}$ and optimizers
            \For{$epoch = 1$ to $E_p$}
                \State Select active clients $U \subseteq N$
                \For{client $i \in U$}
                    \State Load $D_i=\{(x^i_1,x^i_2,y^i_1,y^i_2)\}$
                    \State $(f^i_1,f^i_2)\gets M_c^i(x^i_1,x^i_2)$
                    \State Send $(f^i_1,f^i_2,y^i_1,y^i_2)$ to $e_j$
                \EndFor
                \For{edge server $e_j \in E$ }
                    \State Receive smashed data from connected clients  
                    \State Forward it through edge-side mode
                    \State Train $M_e^j$ (contrastive loss)
                    \State Send back gradient to clients to update
                    \State Forward to $M_{\mathrm{cloud}}$
                \EndFor
                \State Cloud trains $M_{\mathrm{cloud}}$ (task loss)
                \If{$epoch \bmod t_1=0$}
                    \State Edge aggregation: update clients
                \EndIf
                \If{$epoch \bmod t_2=0$}
                    \State Cloud aggregation: update edges
                \EndIf
                \State Evaluate $(M_c,M_e,M_{\mathrm{cloud}})$
            \EndFor
            \State \Return Best weights
        \end{algorithmic}
    \end{algorithm}
    
     In general, our framework presented in Algorithm~\ref{alg:sherlfl} combines semantic-aware model splitting, edge-level representation learning, and hierarchical aggregation to deliver a scalable and communication-efficient solution for federated learning in resource-constrained and heterogeneous environments. By reducing communication volume, optimizing resource utilization, and enabling localized processing, SHeRL-FL not only improves performance but also lowers energy consumption and associated carbon emissions, contributing to more environmentally sustainable AI deployment.

\section{Experiments}
    We evaluated on CIFAR-10, CIFAR-100, and HAM10000 for classification (AlexNet, ResNet-18, ResNet-50) and ISIC-2018 for segmentation (ResNet-50 U-Net). Baselines include FedAvg, FedSGD, FedNova, FedProx, HierFL, SplitFL, and HSFL under IID/non-IID. Classification uses cross-entropy loss with the F1 score while segmentation uses IoU and DICE. The models are trained with Adam ($1\mathrm{e}{-4}$ LR) for fixed communication rounds. Traditional FL/SplitFL simulates 200 clients; HierFL, HSFL, and ours use 200 clients and 10 edges. We measure communication cost (data volume) and computation (avg. client RAM). The experiments were run in PyTorch 2.7.0+cu128, Python 3.9.21, on a NVIDIA RTX 3080 Ti.
    \begin{table*}[!ht]
        \centering
        \begin{tabular}{l|cccccccc|c}
            \hline
            \textbf{Method} &
            \begin{tabular}[c]{@{}c@{}}\textbf{Client}\\ \textbf{smashed}\\ \textbf{upload}\end{tabular} &
            \begin{tabular}[c]{@{}c@{}}\textbf{Edge}\\ \textbf{smashed}\\ \textbf{upload}\end{tabular} &
            \begin{tabular}[c]{@{}c@{}}\textbf{Client}\\ \textbf{model}\\ \textbf{upload}\end{tabular} &
            \begin{tabular}[c]{@{}c@{}}\textbf{Client}\\ \textbf{model}\\ \textbf{download}\end{tabular} &
            \begin{tabular}[c]{@{}c@{}}\textbf{Edge}\\ \textbf{model}\\ \textbf{upload}\end{tabular} &
            \begin{tabular}[c]{@{}c@{}}\textbf{Edge}\\ \textbf{model}\\ \textbf{download}\end{tabular} &
            \begin{tabular}[c]{@{}c@{}}\textbf{Gradient}\\ \textbf{from}\\ \textbf{cloud}\end{tabular} &
            \begin{tabular}[c]{@{}c@{}}\textbf{Gradient}\\ \textbf{from}\\ \textbf{edge}\end{tabular} &
            \textbf{Total (GB)} \\
            \hline
            FedAvg & -- & -- & 351.40 & 3,513.95 & -- & -- & -- & -- & 3,865.35 \\
            FedSGD & -- & -- & 351.40 & 3,513.95 & -- & -- & -- & -- & 3,865.35 \\
            FedNova & -- & -- & 351.40 & 3,513.95 & -- & -- & -- & -- & 3,865.35 \\
            FedProx & -- & -- & 351.40 & 3,513.95 & -- & -- & -- & -- & 3,865.35 \\ \hline
            HierFL & -- & -- & 351.40 & 3,865.34 & 175.70 & -- & -- & -- & 4,392.44 \\ \hline
            SplitFed & 6.87 & -- & 1.09 & 216.82 & -- & -- & 3.73 & -- & 228.51 \\ \hline
            HSFL (A) & 54.93 & 13.74 & 0.22 & 0.93 & 1.34 & 1.34 & 12.22 & 31.25 & 115.97 \\
            HSFL (B) & 54.93 & 13.74 & 0.11 & 0.48 & 0.67 & 0.67 & 12.22 & 31.25 & 114.07 \\
            HSFL (C) & 54.93 & 13.74 & 0.06 & 0.24 & 0.34 & 0.34 & 12.22 & 31.25 & 113.12 \\ \hline
            \textbf{Ours (A)} & \textbf{54.93} & \textbf{13.74} & \textbf{0.22} & \textbf{0.93} & \textbf{1.34} & \textbf{1.34} & -- & \textbf{31.25} & \textbf{103.75} \\
            \textbf{Ours (B)} & \textbf{54.93} & \textbf{13.74} & \textbf{0.11} & \textbf{0.48} & \textbf{0.67} & \textbf{0.67} & -- & \textbf{31.25} & \textbf{101.85} \\
            \textbf{Ours (C)} & \textbf{54.93} & \textbf{13.74} & \textbf{0.06} & \textbf{0.24} & \textbf{0.34} & \textbf{0.34} & -- & \textbf{31.25} & \textbf{100.90} \\
            \hline
        \end{tabular}
        \caption{Communication overhead (GB) on CIFAR-100 with ResNet-18 over 200 rounds. 
        (A) is $(t_1=5, t_2=10)$, (B) is $(t_1=10, t_2=20)$, and (C) is $(t_1=25, t_2=50)$.}
        \label{tab:comm_overhead}
    \end{table*}
        \begin{table}
            \centering
            \begin{tabular}{c|c|c}
                \hline
                \textbf{Methods} & \textbf{IoU (\%)} & \textbf{DICE (\%)} \\ \hline
                FedAvg    & 13.73  & 23.38 \\ 
                FedSGD    & 10.23  & 20.54 \\ 
                FedNova   & 19.77  & 25.62 \\ 
                FedProx   & 22.01  & 29.93 \\ \hline
                SplitFed  & 15.26  & 24.46 \\ \hline
                HierFL    & 15.93  & 25.06 \\ \hline
                HSFL (A)  & 24.45 & 32.20 \\
                HSFL (B)   & 24.12 & 32.05 \\
                HSFL (C)     & 23.88 & 31.74 \\ \hline
                Ours (A)  & 28.30 & 35.10 \\
                Ours (B)   & 27.05 & 34.85 \\
                Ours (C)     & 26.72 & 34.42 \\ \hline
            \end{tabular}
            \caption{Segmentation performance comparison of different FL methods in terms of IoU and DICE scores. (A) is $(t_1=5, t_2=10)$, (B) is $(t_1=10, t_2=20)$, and (C) is $(t_1=25, t_2=50)$.}
            \label{tab:sgmentation}
        \end{table}
        \begin{table*}
            \centering
            \begin{tabular}{l|cc|cc|cc}
                \hline
                \multirow{2}{*}{\textbf{Method}} &
                \multicolumn{2}{c|}{\textbf{CIFAR-10}} &
                \multicolumn{2}{c|}{\textbf{CIFAR-100}} &
                \multicolumn{2}{c}{\textbf{HAM10000}} \\
                \cline{2-7}
                 & \textbf{IID} & \textbf{Non-IID} & \textbf{IID} & \textbf{Non-IID} & \textbf{IID} & \textbf{Non-IID} \\
                \hline
                HSFL (baseline) & 49.18\% & 22.58\% & 22.80\% & 0.20\% & 62.28\% & 24.75\% \\
                Ours (A) & 52.64\% & 23.20\% & 25.47\% & 0.53\% & 65.12\% & 25.41\% \\
                Ours (B) & 57.11\% & 37.55\% & 29.73\% & 3.47\% & 69.38\% & 33.58\% \\
                Ours (C) & 60.50\% & 39.62\% & 32.86\% & 4.46\% & 71.57\% & 35.39\% \\
                Ours (D) & 65.44\% & 45.52\% & 55.23\% & 6.65\% & 78.86\% & 38.36\% \\
                \hline
            \end{tabular}
            \caption{Ablation on CIFAR-10, CIFAR-100, and HAM10000 (IID/Non-IID) with AlexNet, ResNet-18, ResNet-50. HSFL baseline vs SHeRL-FL components in F1 Score at $t_1=5$, $t_2=10$. (A) SHeRL-FL without contrastive loss and role-aware splitting, (B) with contrastive loss, (C) with role-aware splitting, and (D) full configuration.}
            \label{tab:ablation}
        \end{table*}
        \begin{table}[!ht]
            \centering
            \begin{tabular}{l|c|c}
            \hline
            \centering\textbf{Methods} & \textbf{IoU (\%)} & \textbf{DICE (\%)} \\ \hline
            HSFL (baseline) & 24.45 & 32.20 \\
            Ours (A) & 25.18 & 33.04 \\
            Ours (B) & 26.02 & 34.12 \\
            Ours (C) & 26.63 & 34.79 \\
            Ours (D)& 28.30 & 35.10 \\
            \hline
            \end{tabular}
            \caption{Ablation study on ISIC-2018 (IID) with ResNet-50 U-Net, comparing the HSFL baseline and SHeRL-FL variants in IoU and DICE at $t_1=5$ and $t_2=10$. (A) SHeRL-FL without contrastive loss and role-aware splitting, (B) with contrastive loss, (C) with role-aware splitting, and (D) full configuration.}
            \label{tab:ablation_segmentation}
        \end{table}
        \begin{table*}[!ht]
            \centering
            \begin{tabular}{l|cc|cc|cc}
                \hline
                \multirow{2}{*}{\textbf{$m$}} &
                \multicolumn{2}{c|}{\textbf{CIFAR-10}} &
                \multicolumn{2}{c|}{\textbf{CIFAR-100}} &
                \multicolumn{2}{c}{\textbf{HAM10000}} \\ \cline{2-7}
                 & \textbf{IID} & \textbf{Non-IID} & \textbf{IID} & \textbf{Non-IID} & \textbf{IID} & \textbf{Non-IID} \\
                \hline
                0.2 & 54.28\% & 36.00\% & 48.15\% & 25.10\% & 71.42\% & 30.55\% \\
                0.5 & 65.44\% & 45.52\% & 55.23\% & 36.65\% & 78.86\% & 38.36\% \\
                1.0 & 59.77\% & 37.49\% & 52.88\% & 30.12\% & 76.40\% & 34.70\% \\
                1.5 & 60.03\% & 41.99\% & 53.10\% & 32.50\% & 76.85\% & 36.10\% \\
                2.0 & 54.38\% & 40.44\% & 48.65\% & 28.00\% & 72.12\% & 32.45\% \\
                \hline
            \end{tabular}
            \caption{Ablation on CIFAR-10, CIFAR-100, and HAM10000 (IID/Non-IID) with AlexNet, ResNet-18, and ResNet-50. Comparison of different margins $m$ in contrastive loss at edge servers of SHeRL-FL in terms of F1 Score at $t_1=5$, $t_2=10$.}            \label{tab:margin_ablation}
        \end{table*}
        \begin{table}[ht]
            \centering
            \begin{tabular}{l|c|c}
                \hline
                \centering\textbf{$m$} & \textbf{IoU (\%)} & \textbf{DICE (\%)} \\ \hline
                0.2 & 24.50 & 30.80 \\
                0.5 & 28.30 & 35.10 \\
                1.0 & 27.20 & 34.00 \\
                1.5 & 26.80 & 33.60 \\
                2.0 & 25.10 & 31.50 \\
                \hline
            \end{tabular}
            \caption{Ablation on ISIC-2018 (IID) with ResNet-50 U-Net. Comparison of margins $m$ in contrastive loss at edge servers of SHeRL-FL in IoU and DICE at $t_1=5$, $t_2=10$.}
            \label{tab:ablation_segmentation_margin}
        \end{table}
        
     \textbf{F1 Score over training rounds.}
        \begin{figure}[!ht]
            \centering
            \begin{subfigure}[b]{\linewidth}
                \centering
                \includegraphics[width=\linewidth]{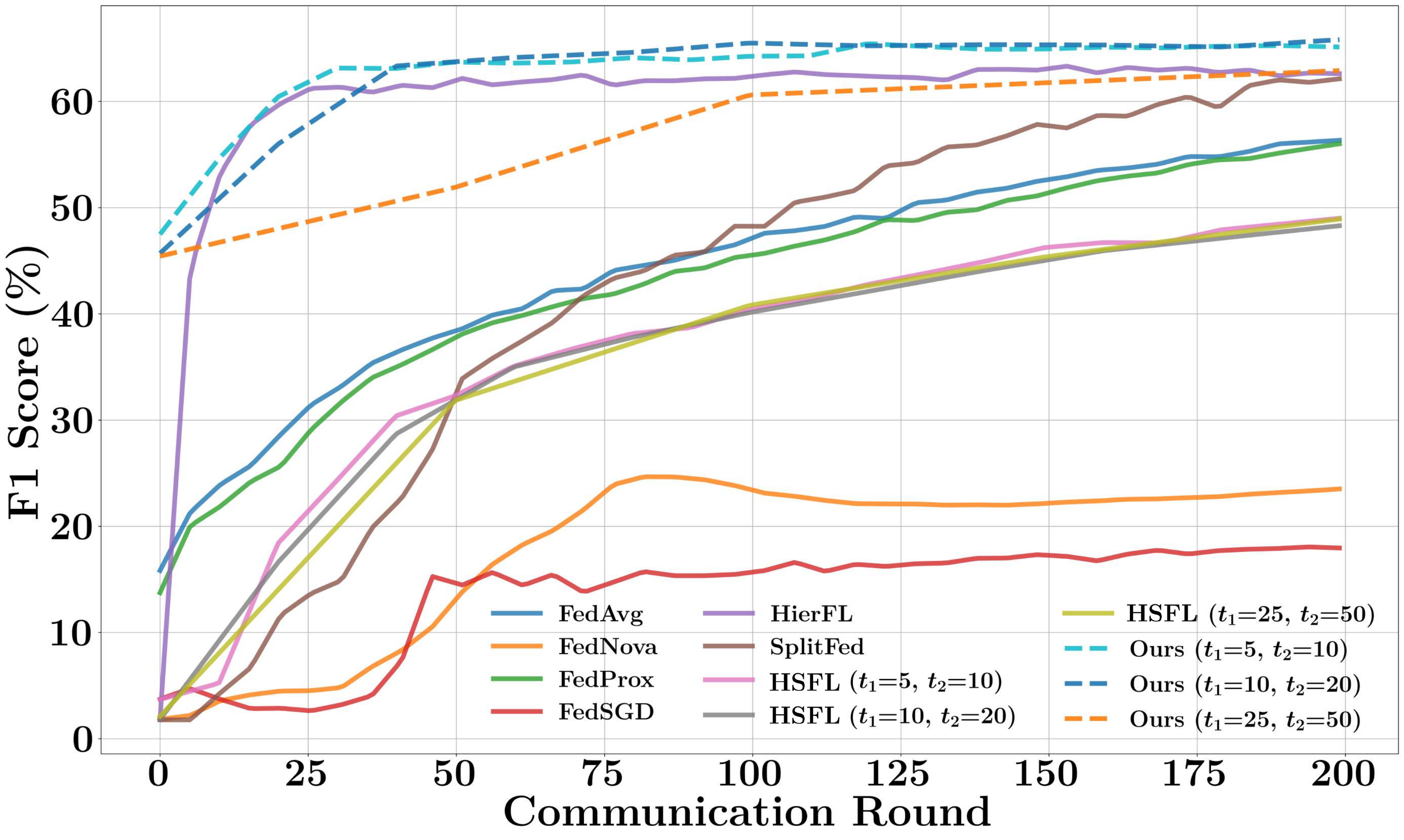}
                \caption{Under IID setting}
                \label{fig:f1_score_per_round_cifar10_iid}
            \end{subfigure}
            \begin{subfigure}[b]{\linewidth}
                \centering
                \includegraphics[width=\linewidth]{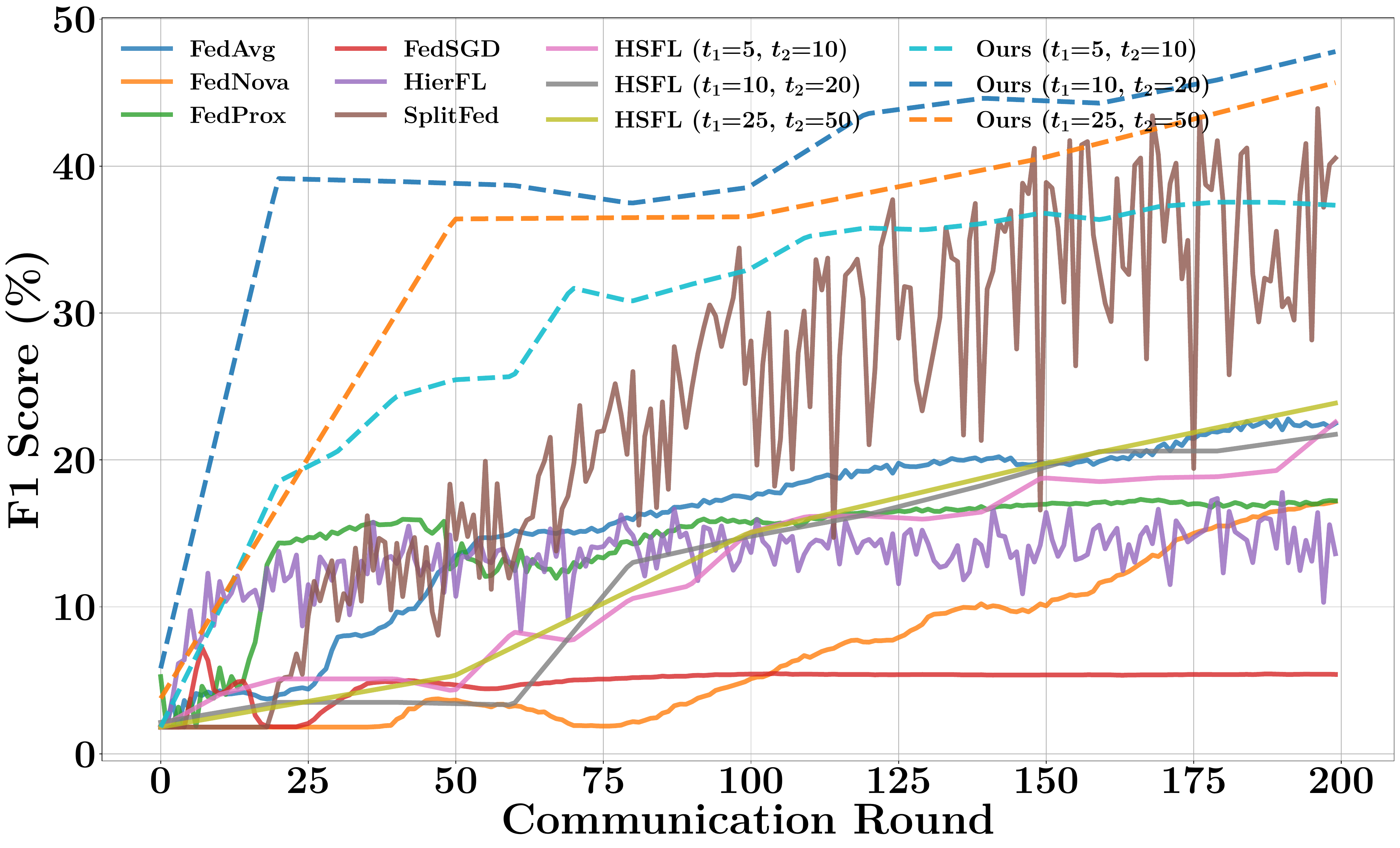}
                \caption{Under non-IID setting}
                \label{fig:f1_score_per_round_cifar10_noniid}
            \end{subfigure}
            \caption{The F1 Score Comparison across methods with the CIFAR-10 dataset and the AlexNet backbone under different settings.}
            \label{fig:f1_score_per_round_cifar10}
        \end{figure}
        \begin{figure}[!ht]
            \centering
            \begin{subfigure}[b]{0.49\linewidth}
                \includegraphics[width=\linewidth]{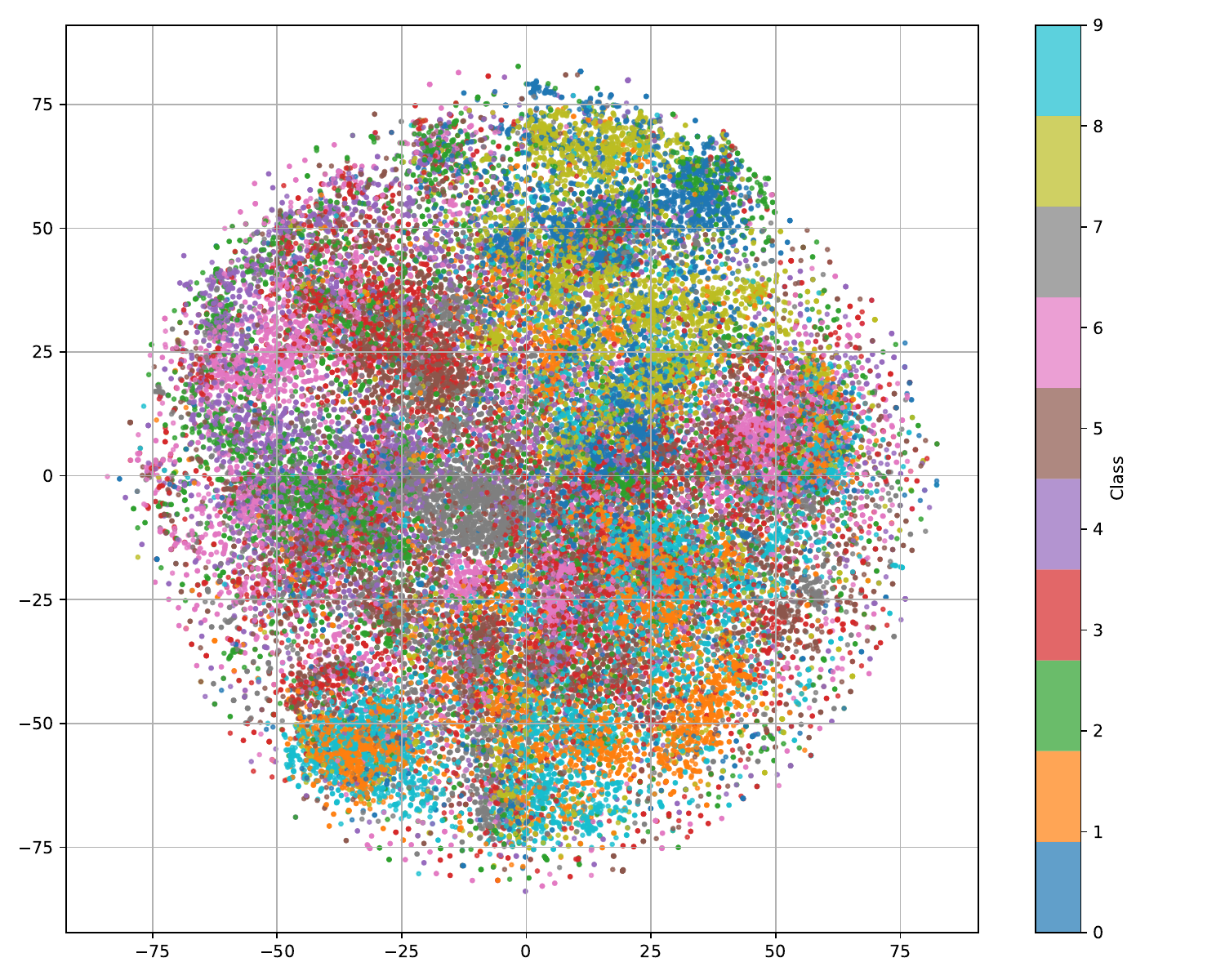}
                \caption{Under IID setting} 
                \label{fig:tsne_encodediid}
            \end{subfigure}
            \hfill
            \begin{subfigure}[b]{0.49\linewidth}
                \includegraphics[width=\linewidth]{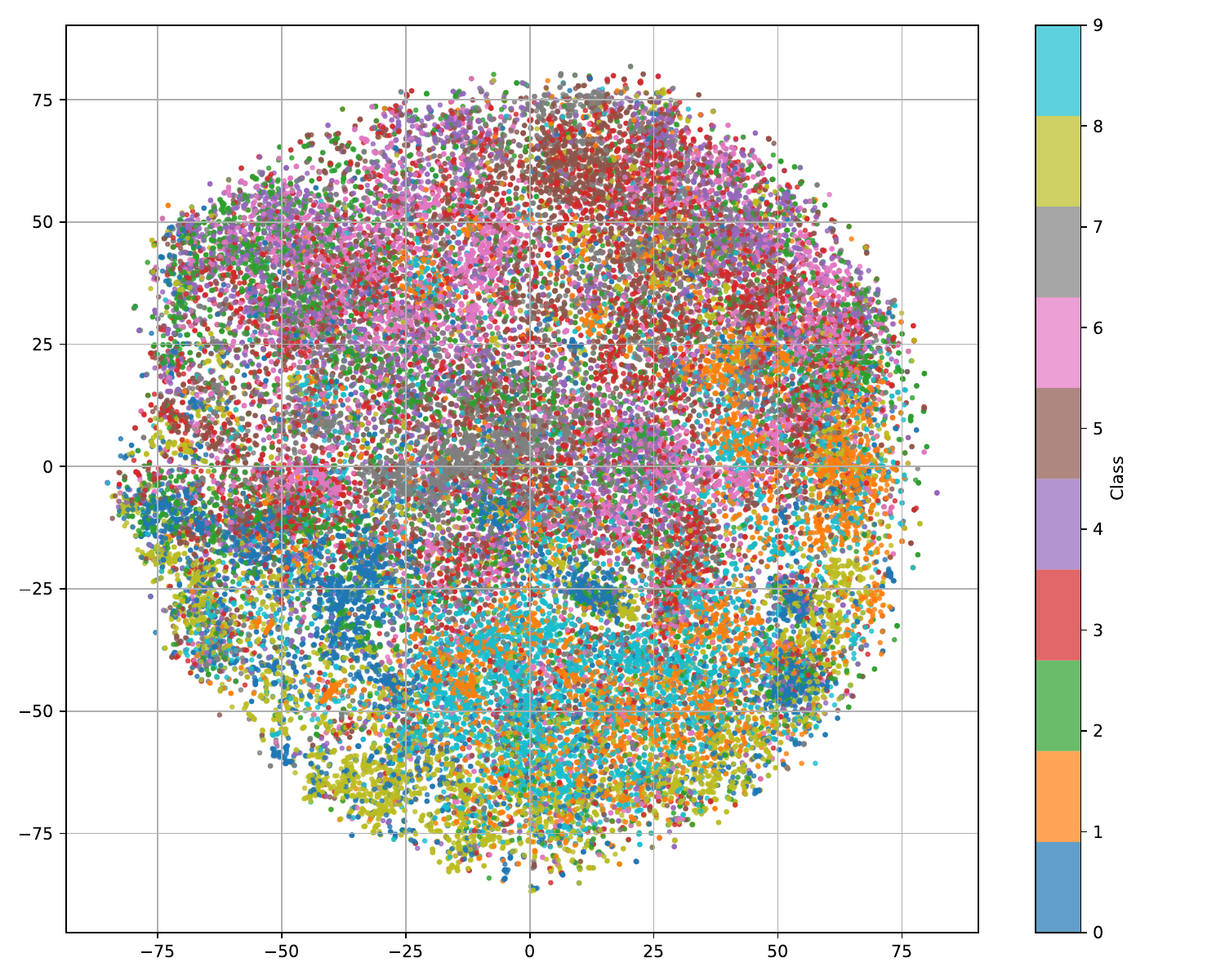}
                \caption{Under non-IID setting} 
                \label{fig:tsne_encodednoniid}
            \end{subfigure}
            \caption{t-SNE visualizations of learned representations of the CIFAR-10 dataset under different settings.}
            \label{fig:tsne_combined}
        \end{figure}
        \begin{figure}[!ht]
            \centering
            \begin{subfigure}[b]{\linewidth}
                \centering
                \includegraphics[width=\linewidth]{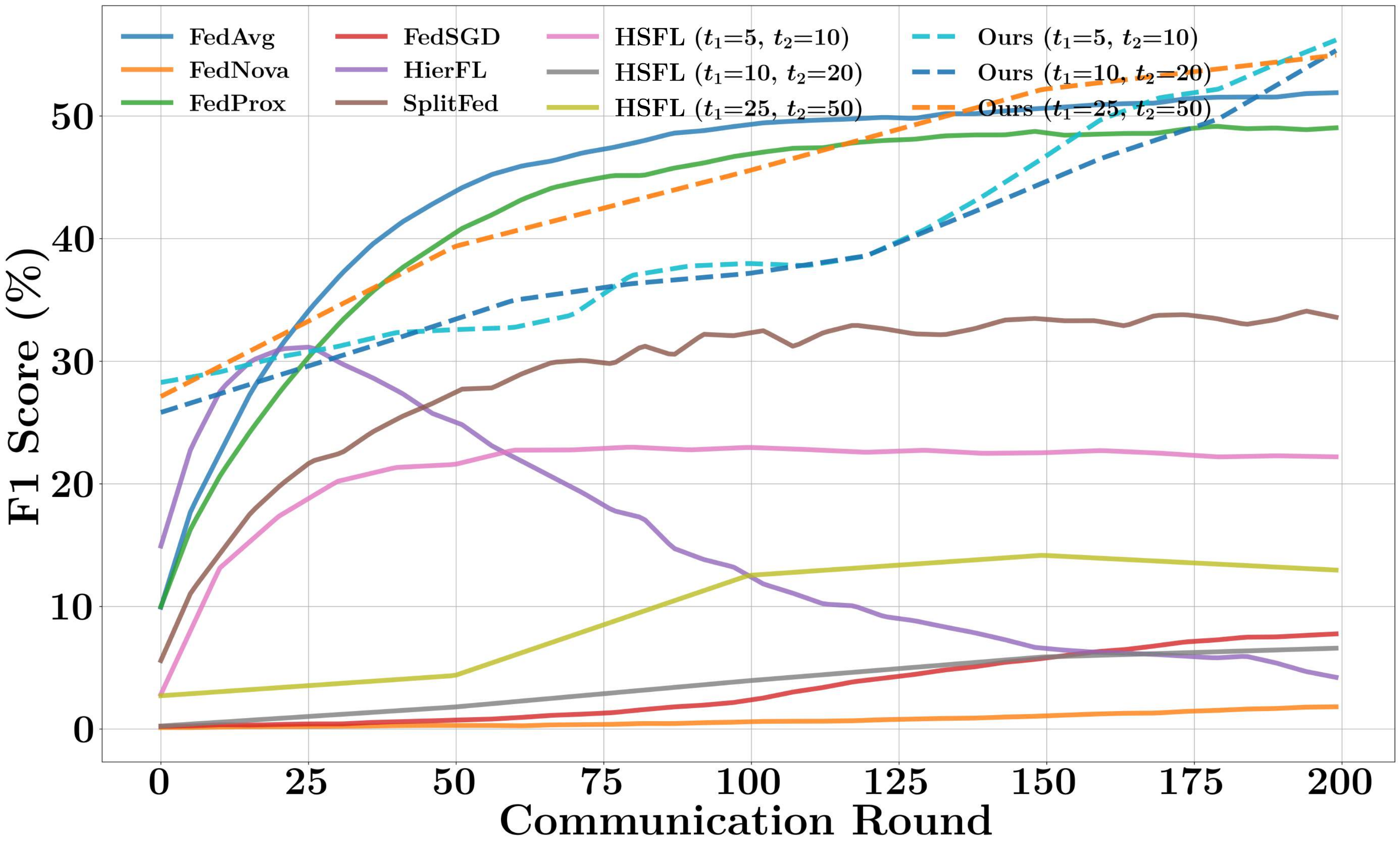}
                \caption{Under IID setting}
                \label{fig:f1_score_per_round_cifar100_iid}
            \end{subfigure}
            \begin{subfigure}[b]{\linewidth}
                \centering
                \includegraphics[width=\linewidth]{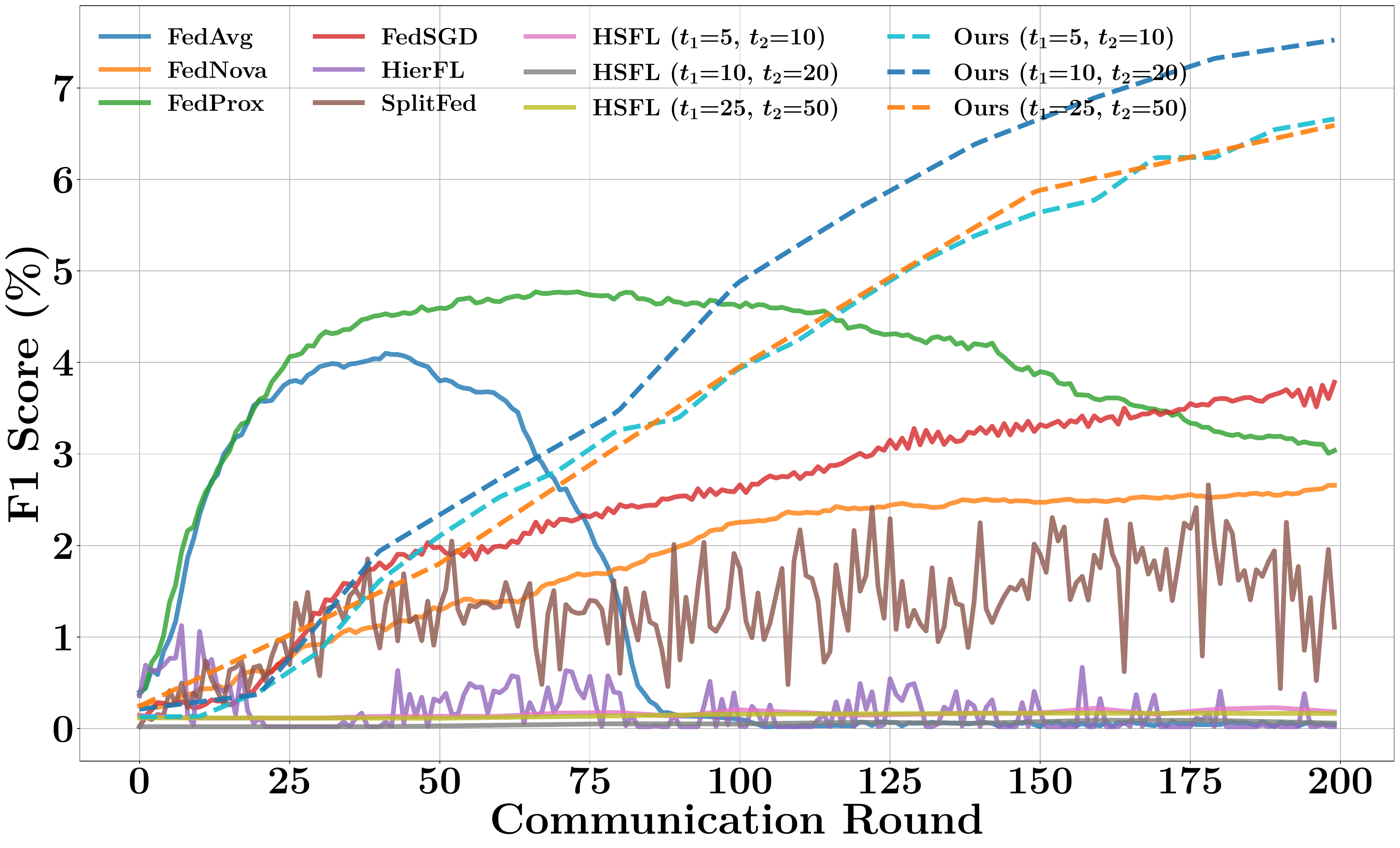}
                \caption{Under non-IID setting}
                \label{fig:f1_score_per_round_cifar100_noniid}
            \end{subfigure}
            \caption{The F1 Score Comparison across methods with the CIFAR-100 dataset and the ResNet-18 backbone under different settings.}
            \label{fig:f1_score_per_round_cifar100} 
        \end{figure}
        \begin{figure}[!ht]
            \centering
            \begin{subfigure}[b]{\linewidth}
                \centering
                \includegraphics[width=\linewidth]{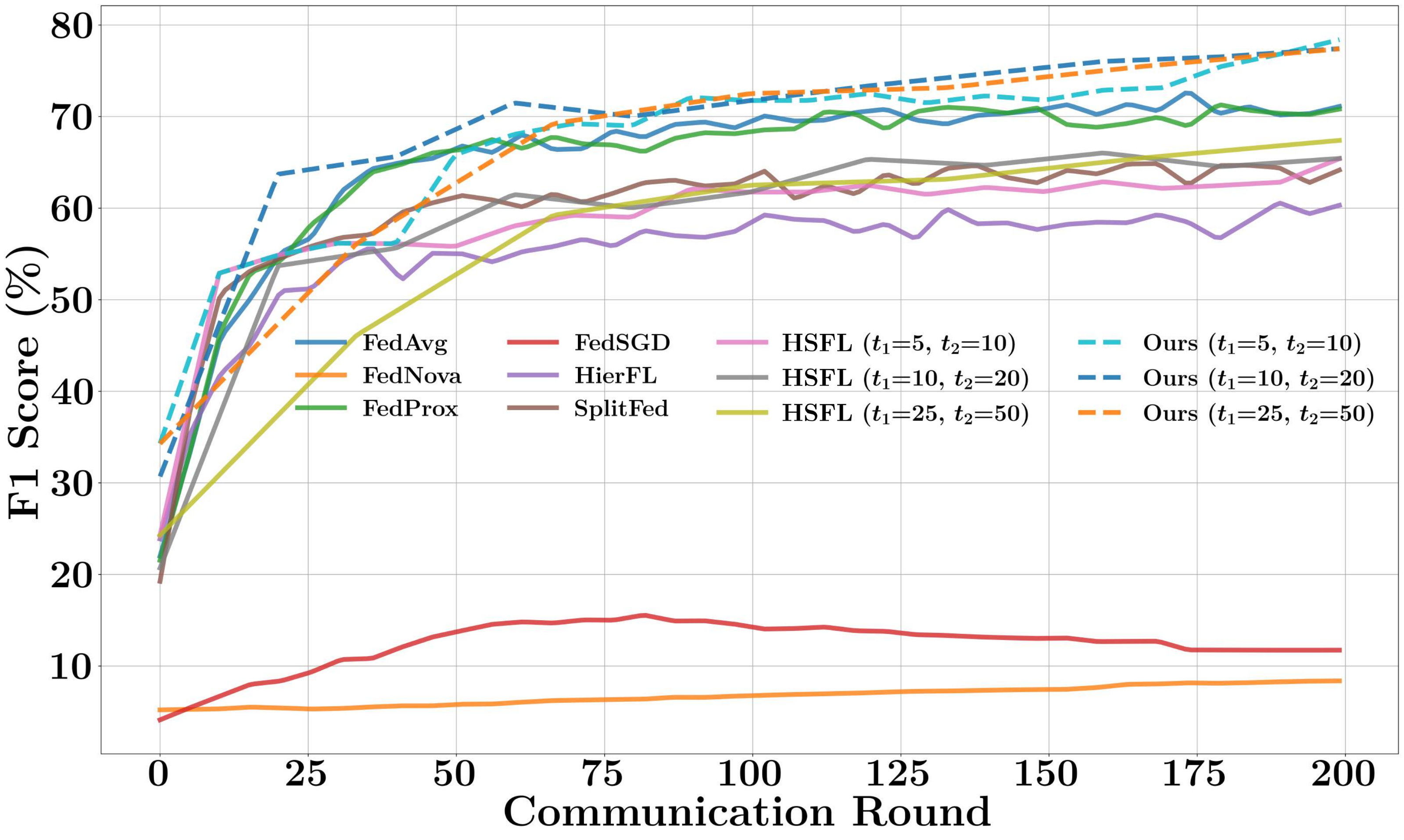}
                \caption{Under IID setting}
                \label{fig:f1_score_per_round_ham10000_iid}
            \end{subfigure}
            \begin{subfigure}[b]{\linewidth}
                \centering
                \includegraphics[width=\linewidth]{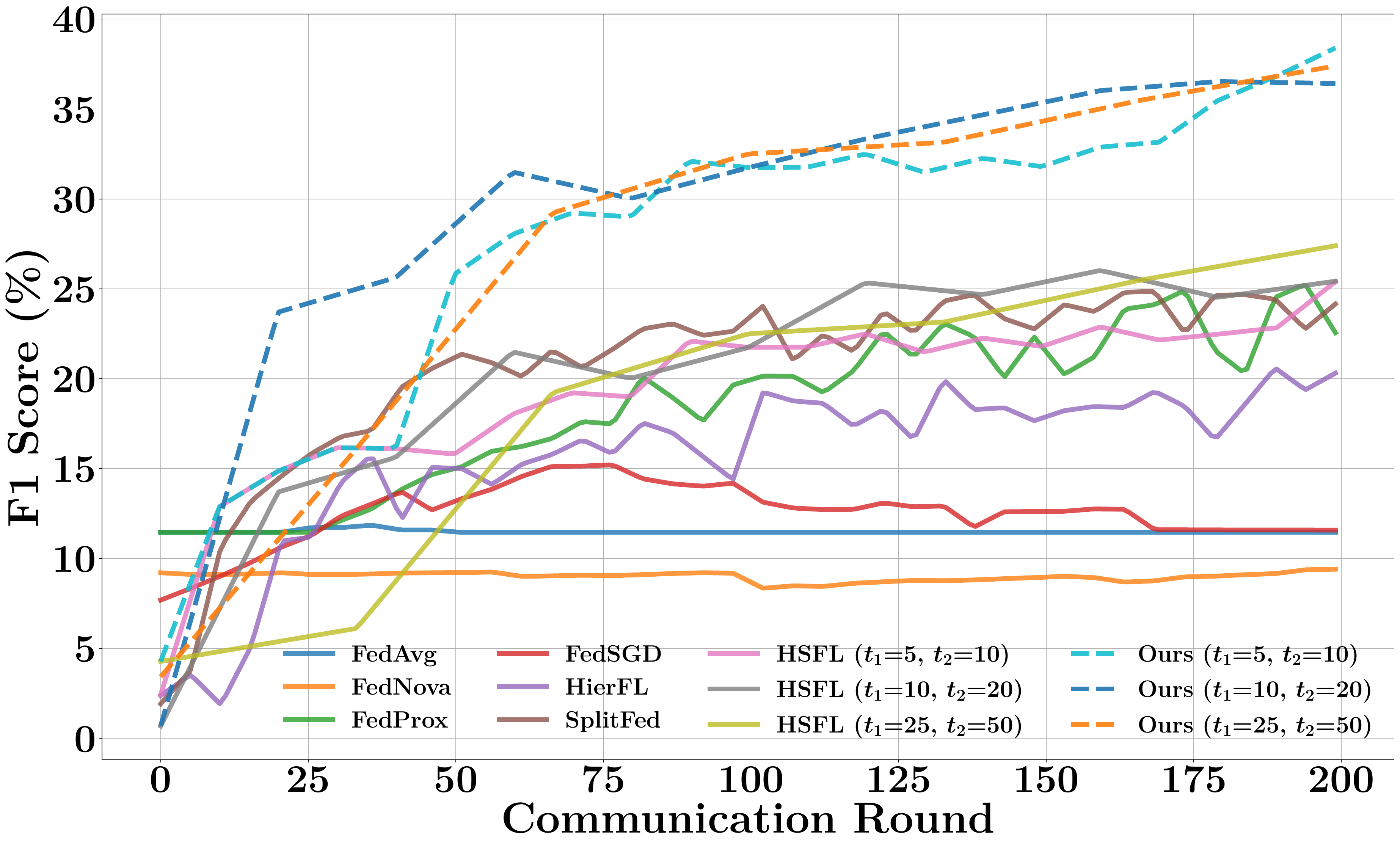}
                \caption{Under non-IID setting}
                \label{fig:f1_score_per_round_ham10000_noniid}
            \end{subfigure}
            \caption{The F1 Score Comparison across methods with the HAM10000 dataset and the ResNet-50 backbone under different settings.}
            \label{fig:f1_score_per_round_ham10000}
        \end{figure}
        In the case of the CIFAR-10 dataset under the IID setting with the AlexNet backbone, Figure~\ref{fig:f1_score_per_round_cifar10}(\subref{fig:f1_score_per_round_cifar10_iid}) shows that our proposed method significantly outperforms all baselines in terms of the F1 score in 200 communication rounds. Specifically, the configuration with synchronization intervals ($t_1 = 10, t_2 = 20$) achieves the highest F1 score, exceeding 65\%. In contrast, traditional methods such as FedAvg, FedSGD, FedNova, and FedProx remain below 55\%, while SplitFed and hierarchical baselines such as HierFL and HSFL converge more slowly and plateau around 60\%. Our method maintains competitive performance even with larger aggregation intervals (e.g., $t_1 = 25, t_2 = 50$). 
        
        In terms of non-IID settings, Figure \ref{fig:f1_score_per_round_cifar10}(\subref{fig:f1_score_per_round_cifar10_noniid}) illustrates our proposed method consistently outperforms all baselines, with the configuration $(t_1=10, t_2=20)$ achieving the highest F1 score. In addition, traditional methods such as FedAvg, FedProx, and FedSGD perform poorly, with slow or stagnant improvements, while SplitFed shows fast initial gains but suffers from instability.
        
        To explain these performances, Figure \ref{fig:tsne_combined} shows the cluster of smashed data from the test CIFAR-10 dataset after going through the edge-side model in IID and non-IID settings. The label follows the distribution of all samples, so the performance of the specific task module in the cloud remains.
        This result highlights the effectiveness and stability of our proposed framework, particularly in scenarios with balanced data distributions and lightweight model architectures.
        
        In both IID (Figure~\ref{fig:f1_score_per_round_cifar100}(\subref{fig:f1_score_per_round_cifar100_iid})) and Non-IID (Figure~\ref{fig:f1_score_per_round_cifar100}(\subref{fig:f1_score_per_round_cifar100_noniid})) settings on CIFAR-100 with ResNet-18, SHeRL-FL consistently outperforms all baselines in F1 score. While FedAvg and FedProx perform reasonably in IID, they degrade sharply in Non-IID. FedNova and FedSGD improve steadily but plateau at lower scores. SplitFed shows instability with large fluctuations, especially in Non-IID. HierFL and all HSPL variants remain near zero in both cases. SHeRL-FL achieves the highest and most stable F1 across rounds, with the best result at $(t_1=10, t_2=20)$, demonstrating robustness to data heterogeneity.
        
        As shown in Figure \ref{fig:f1_score_per_round_ham10000}(\subref{fig:f1_score_per_round_ham10000_iid}) displays the F1 Score in the HAM10000 dataset with IID setting. Specifically, FedAvg and FedProx converge to F1 scores around 70\%. In contrast, SplitFed and HierFL exhibit slower convergence and slightly lower final performance. In addition, HSFL variants show improved results as synchronization intervals ($t_1$, $t_2$) decrease. In particular, the proposed method outperforms all baselines in all configurations, achieving the highest F1 score of approximately 78\% with ($t_1=5$, $ t_2=10$).

        Figure \ref{fig:f1_score_per_round_ham10000}(\subref{fig:f1_score_per_round_ham10000_iid}) presents the results under non-IID conditions. In this setting, the performance of all methods declines relative to the IID scenario. FedSGD exhibits limited improvement over communication rounds, while FedAvg, FedNova, and FedProx show slower and less stable convergence. Furthermore, although HSFL variants benefit from increased synchronization intervals, their performance remains sensitive to data heterogeneity. In comparison, the proposed method consistently achieves the highest F1 scores, with the configuration ($t_1=10$, $t_2=20$) providing the most favorable balance between convergence speed and accuracy.

     \textbf{Performance of Segmentation task.} Table~\ref{tab:sgmentation} presents the segmentation performance of various baselines of FL and our proposed method, evaluated using IoU and DICE scores. Among traditional FL methods, FedProx achieves the highest accuracy, while SplitFed and HierFL offer marginal improvements due to model splitting and hierarchical model aggregation. HSFL demonstrates significant gains over these baselines, while our method consistently outperforms all compared approaches in all aggregation settings, achieving the best results with $t_1{=}25$ and $t_2{=}50$, highlighting the effectiveness of joint split hierarchy representation learning under low frequency synchronization.
        
\begin{figure}[!t]
            \centering
            \includegraphics[width=\linewidth]{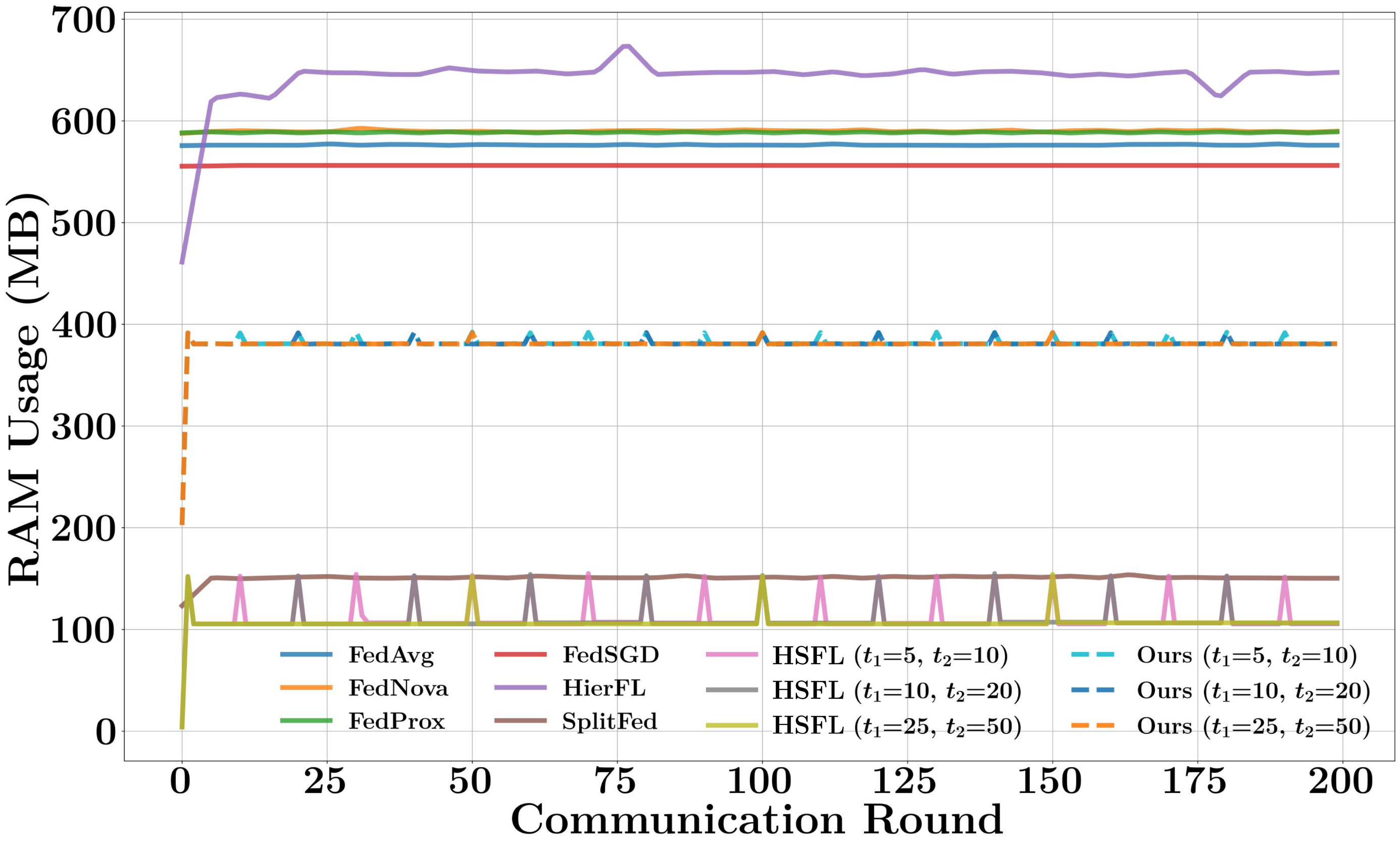}
            \caption{The RAM Usage at client tier across mothods with the CIFAR-10 dataset and the AlexNet backbone under IID Scenario.}
            \label{fig:ram_per_round}
        \end{figure}

     \textbf{Computational cost.} Figure~\ref{fig:ram_per_round} compares average client RAM usage across methods. Our approach maintains stable consumption around 380 MB, well below HierFL ($\sim$650 MB) and flat FL methods such as FedAvg, FedSGD, FedProx, and FedNova ($\sim$580 MB). While SplitFed and HSFL use less memory ($<$150 MB), their accuracy is notably lower. The slight increase over HSFL stems from processing image pairs instead of single images. Overall, our method offers a favorable trade-off, combining strong performance with low, stable memory usage, which is suitable for memory-constrained clients.

     \textbf{Communication cost.} Table~\ref{tab:comm_overhead} reports the communication overhead (GB) for 200 clients. Hierarchical methods assume 200 clients linked through 10 edge servers, while flat baselines communicate directly with the cloud. All methods use a 10\% client sampling rate. Flat methods (FedAvg, FedSGD, FedNova, FedProx) incur very high costs ($>$3{,}865 GB) due to full model exchange each round. HierFL reaches the highest total (4{,}392.44 GB) but shifts much of the load from the cloud to edges, benefiting scenarios with limited cloud bandwidth. SplitFed lowers cost (216.82 GB) by uploading smashed data, but still needs gradient feedback from the cloud. HSFL reduces the total to $\sim$115 GB by limiting gradient exchanges, though cloud-to-edge gradients remain. Our method achieves the lowest overhead (100.90 GB) by removing cloud-to-edge gradient transfers; edges coordinate client updates locally. While smashed activations from clients (54.93 GB) and edges (13.74 GB) add volume, these are transmitted incrementally, minimizing peak bandwidth. Overall, the approach offers strong scalability, bandwidth efficiency, and practicality for large-scale deployments with constrained cloud capacity.
        
     \textbf{Ablation Study.} The ablation results (Tables~\ref{tab:ablation}--\ref{tab:ablation_segmentation}) show that both contrastive loss and role-aware splitting are key to SHeRL-FL’s effectiveness. Removing both components offers only marginal gains over HSFL, while adding either component individually provides noticeable improvements, especially under non-IID settings. The best performance is consistently achieved when both are combined, confirming their complementary roles in improving representation alignment, optimizing partitioning, and improving generalization across tasks.

     The margin study (Tables~\ref{tab:margin_ablation}--\ref{tab:ablation_segmentation_margin}) further reveals that a moderate value of $m=0.5$ produces the most stable and accurate results for classification and segmentation. Smaller margins lead to insufficient interclass separation, whereas larger margins can overly penalize feature alignment, reducing model stability. The consistent peak at $m=0.5$ highlights the importance of carefully tuning the contrast loss margin to balance the strength of separation and the stability of optimization.

\section{Comparison with HSFL}
    Compared to HSFL, SHeRL-FL introduces key architectural and training enhancements that lead to both accuracy and efficiency gains. The integration of contrastive learning at the edge aligns feature spaces without requiring end-to-end gradient flow, reducing representation drift and improving performance, particularly in non-IID settings. Role-aware splitting assigns model segments to tiers more systematically, stabilizing training and accelerating convergence. Communication efficiency is improved by eliminating cloud-to-edge gradient transmission, while maintaining comparable or lower bandwidth usage. Although HSFL achieves lower peak memory usage by processing single images, SHeRL-FL balances slightly higher memory consumption with consistently superior accuracy, faster convergence, and more flexible synchronization. These advantages make SHeRL-FL a more practical choice for large-scale, heterogeneous federated deployments.

\section{Limitations}
    The current role‑aware split is fixed and heuristic; adaptive partitioning could improve efficiency in varied environments. Contrastive learning assumes labeled edge data, which may limit applicability; self‑supervised alternatives are a potential extension. While communication volume is reduced, real network latency tests would further strengthen deployment practicality.
\section{Conclusion}
SHeRL-FL enhances representation learning flexibility while delivering practical communication and memory gains in federated settings. By embedding contrastive learning at the edge, it reduces reliance on cloud synchronization, enabling asynchronous updates and lowering idle time, particularly in heterogeneous or bandwidth-limited environments. Although the communication reduction over HSFL is modest, SHeRL-FL supports more flexible scheduling under real-world constraints. Experiments show consistent accuracy improvements in both IID and non-IID settings, with the largest gains in non-IID cases due to contrastive loss aligning feature spaces. Role-aware splitting further stabilizes tier training and mitigates representation drift, leading to faster convergence. The removal of cloud-to-edge gradient transfers improves bandwidth efficiency and allows more adaptive synchronization, making SHeRL-FL well suited for large-scale, resource-constrained deployments.

\bibliography{aaai2026}

\begin{thebibliography}{30}
\providecommand{\natexlab}[1]{#1}

\bibitem[{Abuadbba et~al.(2020)Abuadbba, Kim, Kim, Thapa, Camtepe, Gao, Kim, and Nepal}]{10.1145/3320269.3384740}
Abuadbba, S.; Kim, K.; Kim, M.; Thapa, C.; Camtepe, S.~A.; Gao, Y.; Kim, H.; and Nepal, S. 2020.
\newblock Can We Use Split Learning on 1D CNN Models for Privacy Preserving Training?
\newblock In \emph{Proceedings of the 15th ACM Asia Conference on Computer and Communications Security}, ASIA CCS '20, 305–318. New York, NY, USA: Association for Computing Machinery.
\newblock ISBN 9781450367509.

\bibitem[{Albuquerque et~al.(2024)Albuquerque, Dias, Ziazet, Vandikas, Ickin, Jaumard, Natalino, Wosinska, Monti, and Wong}]{10707222}
Albuquerque, R.~A.; Dias, L.~P.; Ziazet, M.; Vandikas, K.; Ickin, S.; Jaumard, B.; Natalino, C.; Wosinska, L.; Monti, P.; and Wong, E. 2024.
\newblock { Asynchronous Federated Split Learning }.
\newblock In \emph{2024 IEEE 8th International Conference on Fog and Edge Computing (ICFEC)}, 11--18. Los Alamitos, CA, USA: IEEE Computer Society.

\bibitem[{Chen et~al.(2020)Chen, Kornblith, Norouzi, and Hinton}]{10.5555/3524938.3525087}
Chen, T.; Kornblith, S.; Norouzi, M.; and Hinton, G. 2020.
\newblock A simple framework for contrastive learning of visual representations.
\newblock In \emph{Proceedings of the 37th International Conference on Machine Learning}, ICML'20. JMLR.org.

\bibitem[{Codella et~al.(2018)Codella, Gutman, Celebi, Helba, Marchetti, Dusza, Kalloo, Liopyris, Mishra, Kittler, and Halpern}]{8363547}
Codella, N. C.~F.; Gutman, D.; Celebi, M.~E.; Helba, B.; Marchetti, M.~A.; Dusza, S.~W.; Kalloo, A.; Liopyris, K.; Mishra, N.; Kittler, H.; and Halpern, A. 2018.
\newblock Skin lesion analysis toward melanoma detection: A challenge at the 2017 International symposium on biomedical imaging (ISBI), hosted by the international skin imaging collaboration (ISIC).
\newblock In \emph{2018 IEEE 15th International Symposium on Biomedical Imaging (ISBI 2018)}, 168--172.

\bibitem[{Guo et~al.(2025)Guo, Wang, Zeng, Zhu, Jiang, Wang, Zhou, Wang, Xiong, and Qu}]{guo2025exploringvulnerabilitiesfederatedlearning}
Guo, P.; Wang, R.; Zeng, S.; Zhu, J.; Jiang, H.; Wang, Y.; Zhou, Y.; Wang, F.; Xiong, H.; and Qu, L. 2025.
\newblock Exploring the Vulnerabilities of Federated Learning: A Deep Dive into Gradient Inversion Attacks.
\newblock arXiv:2503.11514.

\bibitem[{Gupta and Raskar(2018)}]{GUPTA20181}
Gupta, O.; and Raskar, R. 2018.
\newblock Distributed learning of deep neural network over multiple agents.
\newblock \emph{Journal of Network and Computer Applications}, 116: 1--8.

\bibitem[{Hatamizadeh et~al.(2023)Hatamizadeh, Yin, Molchanov, Myronenko, Li, Dogra, Feng, Flores, Kautz, Xu, and Roth}]{10025466}
Hatamizadeh, A.; Yin, H.; Molchanov, P.; Myronenko, A.; Li, W.; Dogra, P.; Feng, A.; Flores, M.~G.; Kautz, J.; Xu, D.; and Roth, H.~R. 2023.
\newblock Do Gradient Inversion Attacks Make Federated Learning Unsafe?
\newblock \emph{IEEE Transactions on Medical Imaging}, 42(7): 2044--2056.

\bibitem[{Khan et~al.(2024)Khan, Guizani, Al-Fuqaha, Hong, Niyato, and Han}]{10251444}
Khan, L.~U.; Guizani, M.; Al-Fuqaha, A.; Hong, C.~S.; Niyato, D.; and Han, Z. 2024.
\newblock A Joint Communication and Learning Framework for Hierarchical Split Federated Learning.
\newblock \emph{IEEE Internet of Things Journal}, 11(1): 268--282.

\bibitem[{Khosla et~al.(2020)Khosla, Teterwak, Wang, Sarna, Tian, Isola, Maschinot, Liu, and Krishnan}]{10.5555/3495724.3497291}
Khosla, P.; Teterwak, P.; Wang, C.; Sarna, A.; Tian, Y.; Isola, P.; Maschinot, A.; Liu, C.; and Krishnan, D. 2020.
\newblock Supervised contrastive learning.
\newblock In \emph{Proceedings of the 34th International Conference on Neural Information Processing Systems}, NIPS '20. Red Hook, NY, USA: Curran Associates Inc.
\newblock ISBN 9781713829546.

\bibitem[{Kirkpatrick et~al.(2017)Kirkpatrick, Pascanu, Rabinowitz, Veness, Desjardins, Rusu, Milan, Quan, Ramalho, Grabska-Barwinska, Hassabis, Clopath, Kumaran, and Hadsell}]{doi:10.1073/pnas.1611835114}
Kirkpatrick, J.; Pascanu, R.; Rabinowitz, N.; Veness, J.; Desjardins, G.; Rusu, A.~A.; Milan, K.; Quan, J.; Ramalho, T.; Grabska-Barwinska, A.; Hassabis, D.; Clopath, C.; Kumaran, D.; and Hadsell, R. 2017.
\newblock Overcoming catastrophic forgetting in neural networks.
\newblock \emph{Proceedings of the National Academy of Sciences}, 114(13): 3521--3526.

\bibitem[{Konečný, McMahan, and Ramage(2015)}]{konečný2015federatedoptimizationdistributedoptimizationdatacenter}
Konečný, J.; McMahan, B.; and Ramage, D. 2015.
\newblock Federated Optimization: Distributed Optimization Beyond the Datacenter.
\newblock arXiv:1511.03575.

\bibitem[{Krizhevsky and Hinton(2009)}]{krizhevsky2009learning}
Krizhevsky, A.; and Hinton, G. 2009.
\newblock Learning multiple layers of features from tiny images.
\newblock Technical report, University of Toronto.

\bibitem[{Li et~al.(2020)Li, Sahu, Zaheer, Sanjabi, Talwalkar, and Smith}]{MLSYS2020_1f5fe839}
Li, T.; Sahu, A.~K.; Zaheer, M.; Sanjabi, M.; Talwalkar, A.; and Smith, V. 2020.
\newblock Federated Optimization in Heterogeneous Networks.
\newblock In Dhillon, I.; Papailiopoulos, D.; and Sze, V., eds., \emph{Proceedings of Machine Learning and Systems}, volume~2, 429--450.

\bibitem[{Li and Lyu(2023)}]{li2023convergence}
Li, Y.; and Lyu, X. 2023.
\newblock Convergence Analysis of Sequential Federated Learning on Heterogeneous Data.
\newblock In \emph{Thirty-seventh Conference on Neural Information Processing Systems}.

\bibitem[{Li, Wang, and An(2023)}]{10.1145/3580795}
Li, Y.; Wang, X.; and An, L. 2023.
\newblock Hierarchical Clustering-based Personalized Federated Learning for Robust and Fair Human Activity Recognition.
\newblock \emph{Proc. ACM Interact. Mob. Wearable Ubiquitous Technol.}, 7(1).

\bibitem[{Lin et~al.(2025)Lin, Wei, Chen, Lam, Chen, Gao, and Luo}]{10980018}
Lin, Z.; Wei, W.; Chen, Z.; Lam, C.-T.; Chen, X.; Gao, Y.; and Luo, J. 2025.
\newblock Hierarchical Split Federated Learning: Convergence Analysis and System Optimization.
\newblock \emph{IEEE Transactions on Mobile Computing}, 1--16.

\bibitem[{Liu et~al.(2020)Liu, Zhang, Song, and Letaief}]{9148862}
Liu, L.; Zhang, J.; Song, S.; and Letaief, K.~B. 2020.
\newblock Client-Edge-Cloud Hierarchical Federated Learning.
\newblock In \emph{ICC 2020 - 2020 IEEE International Conference on Communications (ICC)}, 1--6.

\bibitem[{Mao et~al.(2025)Mao, Liu, Tian, Pan, Trucco, and Lin}]{10746501}
Mao, J.; Liu, J.; Tian, X.; Pan, Y.; Trucco, E.; and Lin, H. 2025.
\newblock Toward Integrating Federated Learning With Split Learning via Spatio-Temporal Graph Framework for Brain Disease Prediction.
\newblock \emph{IEEE Transactions on Medical Imaging}, 44(3): 1334--1346.

\bibitem[{McMahan et~al.(2017)McMahan, Moore, Ramage, Hampson, and y~Arcas}]{mcmahan2017communication}
McMahan, B.; Moore, E.; Ramage, D.; Hampson, S.; and y~Arcas, B.~A. 2017.
\newblock Communication-efficient learning of deep networks from decentralized data.
\newblock In \emph{Artificial intelligence and statistics}, 1273--1282. PMLR.

\bibitem[{Nguyen et~al.(2024)Nguyen, Nguyen, Chang, Pham, Narayanan, and Pazzani}]{Nguyen_2024_CVPR}
Nguyen, H.; Nguyen, H.; Chang, M.; Pham, H.; Narayanan, S.; and Pazzani, M. 2024.
\newblock ConPro: Learning Severity Representation for Medical Images using Contrastive Learning and Preference Optimization.
\newblock In \emph{Proceedings of the IEEE/CVF Conference on Computer Vision and Pattern Recognition (CVPR) Workshops}, 5105--5112.

\bibitem[{Otoum, Guizani, and Mouftah(2023)}]{9756883}
Otoum, S.; Guizani, N.; and Mouftah, H. 2023.
\newblock On the Feasibility of Split Learning, Transfer Learning and Federated Learning for Preserving Security in ITS Systems.
\newblock \emph{IEEE Transactions on Intelligent Transportation Systems}, 24(7): 7462--7470.

\bibitem[{Poirot et~al.(2019)Poirot, Vepakomma, Chang, Kalpathy-Cramer, Gupta, and Raskar}]{poirot2019split}
Poirot, M.~G.; Vepakomma, P.; Chang, K.; Kalpathy-Cramer, J.; Gupta, R.; and Raskar, R. 2019.
\newblock Split Learning for Collaborative Deep Learning in Healthcare.
\newblock In \emph{Proceedings of the Machine Learning for Health (ML4H) Workshop at NeurIPS}. Vancouver, Canada.

\bibitem[{Priyadarshini(2024)}]{bdcc8030021}
Priyadarshini, I. 2024.
\newblock Anomaly Detection of IoT Cyberattacks in Smart Cities Using Federated Learning and Split Learning.
\newblock \emph{Big Data and Cognitive Computing}, 8(3).

\bibitem[{Qin et~al.(2023)Qin, Zhang, Liu, and Qian}]{10.1186/s13677-023-00435-5}
Qin, J.; Zhang, X.; Liu, B.; and Qian, J. 2023.
\newblock A split-federated learning and edge-cloud based efficient and privacy-preserving large-scale item recommendation model.
\newblock \emph{J. Cloud Comput.}, 12(1).

\bibitem[{Thapa et~al.(2022)Thapa, Arachchige, Camtepe, and Sun}]{thapa2022splitfed}
Thapa, C.; Arachchige, P. C.~M.; Camtepe, S.; and Sun, L. 2022.
\newblock Splitfed: When federated learning meets split learning.
\newblock In \emph{Proceedings of the AAAI conference on artificial intelligence}, volume~36, 8485--8493.

\bibitem[{Tran et~al.(2025)Tran, Vu, Tran, Hoang, Pham, Nguyen, and Nguyen}]{10980996}
Tran, D.~T.; Vu, H.; Tran, A.; Hoang, T.; Pham, H.; Nguyen, H.; and Nguyen, N.-P. 2025.
\newblock SEMISE: Semi-Supervised Learning for Severity Representation in Medical Image.
\newblock In \emph{2025 IEEE 22nd International Symposium on Biomedical Imaging (ISBI)}, 1--4.

\bibitem[{Tschandl, Rosendahl, and Kittler(2018)}]{tschandl2018ham10000}
Tschandl, P.; Rosendahl, C.; and Kittler, H. 2018.
\newblock The HAM10000 dataset: A large collection of multi-source dermatoscopic images of common pigmented skin lesions.
\newblock \emph{Scientific Data}, 5: 180161.

\bibitem[{Vepakomma et~al.(2019)Vepakomma, Gupta, Swedish, and Raskar}]{vepakomma2019split}
Vepakomma, P.; Gupta, O.; Swedish, T.; and Raskar, R. 2019.
\newblock Split learning for health: Distributed deep learning without sharing raw patient data.
\newblock In \emph{AI for Social Good Workshop, International Conference on Learning Representations (ICLR)}.

\bibitem[{Wang et~al.(2020)Wang, Liu, Liang, Joshi, and Poor}]{10.5555/3495724.3496362}
Wang, J.; Liu, Q.; Liang, H.; Joshi, G.; and Poor, H.~V. 2020.
\newblock Tackling the objective inconsistency problem in heterogeneous federated optimization.
\newblock In \emph{Proceedings of the 34th International Conference on Neural Information Processing Systems}, NIPS '20. Red Hook, NY, USA: Curran Associates Inc.
\newblock ISBN 9781713829546.

\bibitem[{Xie, Xiong, and Luo(2025)}]{pmlr-v260-xie25c}
Xie, W.; Xiong, R.; and Luo, J. 2025.
\newblock Hierarchical Global Asynchronous Federated Learning Across Multi-Center.
\newblock In Nguyen, V.; and Lin, H.-T., eds., \emph{Proceedings of the 16th Asian Conference on Machine Learning}, volume 260 of \emph{Proceedings of Machine Learning Research}, 543--558. PMLR.

\end{thebibliography}


\end{document}